\documentclass{article} 
\usepackage{iclr2017_conference,times}
\usepackage{url}

\usepackage[english]{babel}
\usepackage[utf8]{inputenc}
\usepackage{fancyhdr,graphicx,amsmath,amssymb,subcaption}
\usepackage[ruled,vlined]{algorithm2e}
\usepackage{multirow}
\usepackage{wrapfig}

\newcommand*{\eg}{e.g.\@\xspace}
\newcommand*{\ie}{i.e.\@\xspace}
\newcommand*{\wrt}{w.r.t.\@\xspace}

\newcommand*{\eq}{Eq.\@\xspace}
\newcommand*{\vs}{vs.\@\xspace}

\newcommand*{\registered}{{\ooalign{\hfil\raise .00ex\hbox{\scriptsize R}\hfil\crcr\mathhexbox20D}}\@\xspace}

\title{Semi-Supervised Classification with \\Graph Convolutional Networks}

\author{Thomas N. Kipf\\
University of Amsterdam \\
\texttt{T.N.Kipf@uva.nl} \\
\And
Max Welling\\
University of Amsterdam \\
Canadian Institute for Advanced Research (CIFAR)\\
\texttt{M.Welling@uva.nl} \\
}

\iclrfinalcopy

\begin{document}

\maketitle

\begin{abstract}
We present a scalable approach for semi-supervised learning on graph-structured data that is based on an efficient variant of convolutional neural networks which operate directly on graphs. We motivate the choice of our convolutional architecture via a localized first-order approximation of spectral graph convolutions. Our model scales linearly in the number of graph edges and learns hidden layer representations that encode both local graph structure and features of nodes. In a number of experiments on citation networks and on a knowledge graph dataset we demonstrate that our approach outperforms related methods by a significant margin.
\end{abstract}

\section{Introduction}
We consider the problem of classifying nodes (such as documents) in a graph (such as a citation network), where labels are only available for a small subset of nodes. This problem can be framed as graph-based semi-supervised learning, where label information is smoothed over the graph via some form of explicit graph-based regularization \citep{zhu2003semi, zhou2004learning, belkin2006manifold, weston2012deep}, \eg by using a graph Laplacian regularization term in the loss function:
\begin{equation}
  \mathcal{L} = \mathcal{L}_0 + \lambda \mathcal{L}_{\text{reg}}\, , \quad \text{with} \quad \mathcal{L}_{\text{reg}}= \sum_{i,j}A_{ij}\Vert f(X_i)-f(X_j) \Vert ^2 = f(X)^\top \Delta f(X) \, .
\label{eq:graph-reg}
\end{equation}
Here, $\mathcal{L}_0$ denotes the supervised loss \wrt the labeled part of the graph, $f(\cdot)$ can be a neural network-like differentiable function, $\lambda$ is a weighing factor and $X$ is a matrix of node feature vectors $X_i$. $\Delta = D - A$ denotes the unnormalized graph Laplacian of an undirected graph $\mathcal{G}=(\mathcal{V}, \mathcal{E})$ with $N$ nodes $v_i \in \mathcal{V}$, edges $(v_i, v_j)\in\mathcal{E}$, an adjacency matrix $A\in\mathbb{R}^{N\times N}$ (binary or weighted) and a degree matrix $D_{ii} = \sum_j A_{ij}$. The formulation of \eq \ref{eq:graph-reg} relies on the assumption that connected nodes in the graph are likely to share the same label. This assumption, however, might restrict modeling capacity, as graph edges need not necessarily encode node similarity, but could contain additional information.

In this work, we encode the graph structure directly using a neural network model $f(X, A)$ and train on a supervised target $\mathcal{L}_0$ for all nodes with labels, thereby avoiding explicit graph-based regularization in the loss function. Conditioning $f(\cdot)$ on the adjacency matrix of the graph will allow the model to distribute gradient information from the supervised loss $\mathcal{L}_0$ and will enable it to learn representations of nodes both with and without labels.

Our contributions are two-fold. Firstly, we introduce a simple and well-behaved layer-wise propagation rule for neural network models which operate directly on graphs and show how it can be motivated from a first-order approximation of spectral graph convolutions \citep{hammond2011wavelets}. Secondly, we demonstrate how this form of a graph-based neural network model can be used for fast and scalable semi-supervised classification of nodes in a graph. Experiments on a number of datasets demonstrate that our model compares favorably both in classification accuracy and efficiency (measured in wall-clock time) against state-of-the-art methods for semi-supervised learning.

\section{Fast Approximate Convolutions on Graphs}
\label{sec:fast-convs}

In this section, we provide theoretical motivation for a specific graph-based neural network model $f(X, A)$ that we will use in the rest of this paper. We consider a multi-layer Graph Convolutional Network (GCN) with the following layer-wise propagation rule:
\begin{equation}
  \textstyle
  H^{(l+1)}= \sigma\!\left(\tilde{D}^{-\frac{1}{2}} \tilde{A}\tilde{D}^{-\frac{1}{2}}H^{(l)} W^{(l)} \right) \, .
\label{eq:gcn-layer}
\end{equation}
Here, $\tilde{A} = A + I_N$ is the adjacency matrix of the undirected graph $\mathcal{G}$ with added self-connections.  $I_N$ is the identity matrix, $\tilde{D}_{ii} = \sum_j \tilde{A}_{ij}$ and $W^{(l)}$ is a layer-specific trainable weight matrix. $\sigma(\cdot)$ denotes an activation function, such as the $\mathrm{ReLU}(\cdot) = \max(0,\cdot)$. $H^{(l)}\in \mathbb{R}^{N\times D}$ is the matrix of activations in the $l^{\text{th}}$ layer; $H^{(0)}=X$. In the following, we show that the form of this propagation rule can be motivated\footnote{We provide an alternative interpretation of this propagation rule based on the Weisfeiler-Lehman algorithm \citep{weisfeiler1968reduction} in Appendix \ref{sec:wl}.} via a first-order approximation of localized spectral filters on graphs \citep{hammond2011wavelets, defferrard2016convolutional}.

\subsection{Spectral Graph Convolutions}
We consider spectral convolutions on graphs defined as the multiplication of a signal $x\in \mathbb{R}^N$ (a scalar for every node) with a filter $g_{\theta}=\text{diag}(\theta)$ parameterized by $\theta\in \mathbb{R}^N$ in the Fourier domain, \ie:
\begin{equation}
  g_{\theta} \star x =  Ug_{\theta}U^\top x \, ,
\label{eq:fourier-conv}
\end{equation}
where $U$ is the matrix of eigenvectors of the normalized graph Laplacian $L = I_N - D^{-\frac{1}{2}}AD^{-\frac{1}{2}} = U\Lambda U^\top$, with a diagonal matrix of its eigenvalues $\Lambda$ and $U^\top x$ being the graph Fourier transform of $x$. We can understand $g_{\theta}$ as a function of the eigenvalues of $L$, \ie $g_{\theta}(\Lambda)$. Evaluating \eq \ref{eq:fourier-conv} is computationally expensive, as multiplication with the eigenvector matrix $U$ is $\mathcal{O}(N^2)$. Furthermore, computing the eigendecomposition of $L$ in the first place might be prohibitively expensive for large graphs. To circumvent this problem, it was suggested in \cite{hammond2011wavelets} that $g_{\theta}(\Lambda)$ can be well-approximated by a truncated expansion in terms of Chebyshev polynomials $T_k(x)$ up to $K^{\text{th}}$ order:
\begin{equation}
  g_{\theta'}(\Lambda) \approx \sum_{k=0}^{K} \theta_k ' T_k(\tilde{\Lambda}) \, ,
\label{eq:tchebyshew}
\end{equation}
with a rescaled $\tilde{\Lambda}=\frac{2}{\lambda_{\text{max}}}\Lambda-I_N$. $\lambda_{\text{max}}$ denotes the largest eigenvalue of $L$. $\theta'\in \mathbb{R}^K$ is now a vector of Chebyshev coefficients. The Chebyshev polynomials are recursively defined as $T_k(x) = 2xT_{k-1}(x) - T_{k-2}(x)$, with $T_0(x)=1$ and $T_1(x)=x$. The reader is referred to \cite{hammond2011wavelets} for an in-depth discussion of this approximation.

Going back to our definition of a convolution of a signal $x$ with a filter $g_{\theta'}$, we now have:
\begin{equation}
  g_{\theta'} \star x \approx  \sum_{k=0}^{K} \theta_k' T_k(\tilde{L}) x \, ,
\label{eq:fourier-conv-approx}
\end{equation}
with $\tilde{L} = \frac{2}{\lambda_{\text{max}}}L-I_N$; as can easily be verified by noticing that $(U\Lambda U^\top)^k = U \Lambda^k U^\top$. Note that this expression is now $K$-localized since it is a $K^{\text{th}}$-order polynomial in the Laplacian, \ie it depends only on nodes that are at maximum $K$ steps away from the central node ($K^{\text{th}}$-order neighborhood). The complexity of evaluating \eq \ref{eq:fourier-conv-approx} is $\mathcal{O}(|\mathcal{E}|)$, \ie linear in the number of edges. \cite{defferrard2016convolutional} use this $K$-localized convolution to define a convolutional neural network on graphs.

\subsection{Layer-Wise Linear Model}
\label{sec:linear-model}
A neural network model based on graph convolutions can therefore be built by stacking multiple convolutional layers of the form of \eq \ref{eq:fourier-conv-approx}, each layer followed by a point-wise non-linearity. Now, imagine we limited the layer-wise convolution operation to $K=1$ (see \eq \ref{eq:fourier-conv-approx}), \ie a function that is linear \wrt $L$ and therefore a linear function on the graph Laplacian spectrum.

In this way, we can still recover a rich class of convolutional filter functions by stacking multiple such layers, but we are not limited to the explicit parameterization given by, \eg, the Chebyshev polynomials. We intuitively expect that such a model can alleviate the problem of overfitting on local neighborhood structures for graphs with very wide node degree distributions, such as social networks, citation networks, knowledge graphs and many other real-world graph datasets. Additionally, for a fixed computational budget, this layer-wise linear formulation allows us to build deeper models, a practice that is known to improve modeling capacity on a number of domains \citep{he2015deep}.

In this linear formulation of a GCN we further approximate $\lambda_{\text{max}}\approx 2$, as we can expect that neural network parameters will adapt to this change in scale during training. Under these approximations \eq \ref{eq:fourier-conv-approx} simplifies to:
\begin{equation}
  g_{\theta'} \star x \approx  \theta_0' x
  + \theta_1' \left(L-I_N\right)x = \theta_0' x - \theta_1' D^{-\frac{1}{2}}AD^{-\frac{1}{2}} x \, ,
\label{eq:fourier-conv-approx2}
\end{equation}

with two free parameters $\theta_0'$ and $\theta_1'$. The filter parameters can be shared over the whole graph. Successive application of filters of this form then effectively convolve the $k^{\text{th}}$-order neighborhood of a node, where $k$ is the number of successive filtering operations or convolutional layers in the neural network model.

In practice, it can be beneficial to constrain the number of parameters further to address overfitting and to minimize the number of operations (such as matrix multiplications) per layer. This leaves us with the following expression:
\begin{equation}
  g_{\theta} \star x \approx  \theta \left(I_N + D^{-\frac{1}{2}}AD^{-\frac{1}{2}}\right) x \, ,
\label{eq:fourier-conv-approx3}
\end{equation}
with a single parameter $\theta = \theta_0'=-\theta_1'$. Note that $I_N + D^{-\frac{1}{2}}AD^{-\frac{1}{2}}$ now has eigenvalues in the range $[0, 2]$. Repeated application of this operator can therefore lead to numerical instabilities and exploding/vanishing gradients when used in a deep neural network model. To alleviate this problem, we introduce the following \textit{renormalization trick}: $I_N + D^{-\frac{1}{2}}AD^{-\frac{1}{2}}\rightarrow \tilde{D}^{-\frac{1}{2}}\tilde{A}\tilde{D}^{-\frac{1}{2}}$, with $\tilde{A} = A + I_N$ and $\tilde{D}_{ii} = \sum_j \tilde{A}_{ij}$.

We can generalize this definition to a signal $X\in\mathbb{R}^{N\times C}$ with $C$ input channels (\ie a $C$-dimensional feature vector for every node) and $F$ filters or feature maps as follows:
\begin{equation}
  Z = \tilde{D}^{-\frac{1}{2}}\tilde{A}\tilde{D}^{-\frac{1}{2}}X\Theta \, ,
\label{eq:fourier-conv-approx4}
\end{equation}
where $\Theta\in\mathbb{R}^{C\times F}$ is now a matrix of filter parameters and $Z\in\mathbb{R}^{N\times F}$ is the convolved signal matrix. This filtering operation has complexity $\mathcal{O}(|\mathcal{E}|FC)$, as $\tilde{A}X$ can be efficiently implemented as a product of a sparse matrix with a dense matrix.

\section{Semi-Supervised Node Classification}
\begin{figure}[htp]
    \centering
    \begin{subfigure}[b]{0.67\textwidth}
        \centering
    \includegraphics[clip, trim=5cm 18.8cm 6.4cm 4.4cm, width=\textwidth]{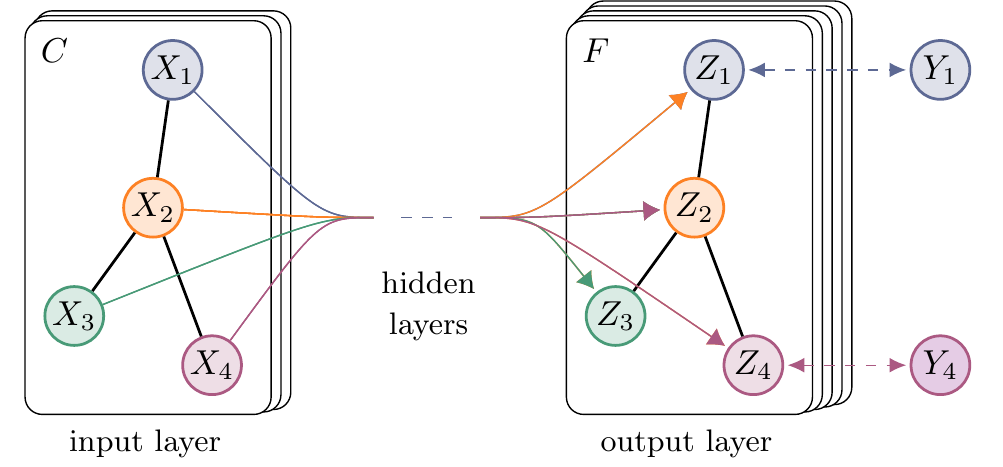}
        \caption{Graph Convolutional Network}
        \label{fig:model-a}
    \end{subfigure}%
    ~
    \begin{subfigure}[b]{0.33\textwidth}
        \centering
        \includegraphics[width=\textwidth, trim={4cm 4cm 4cm 4cm}, clip]{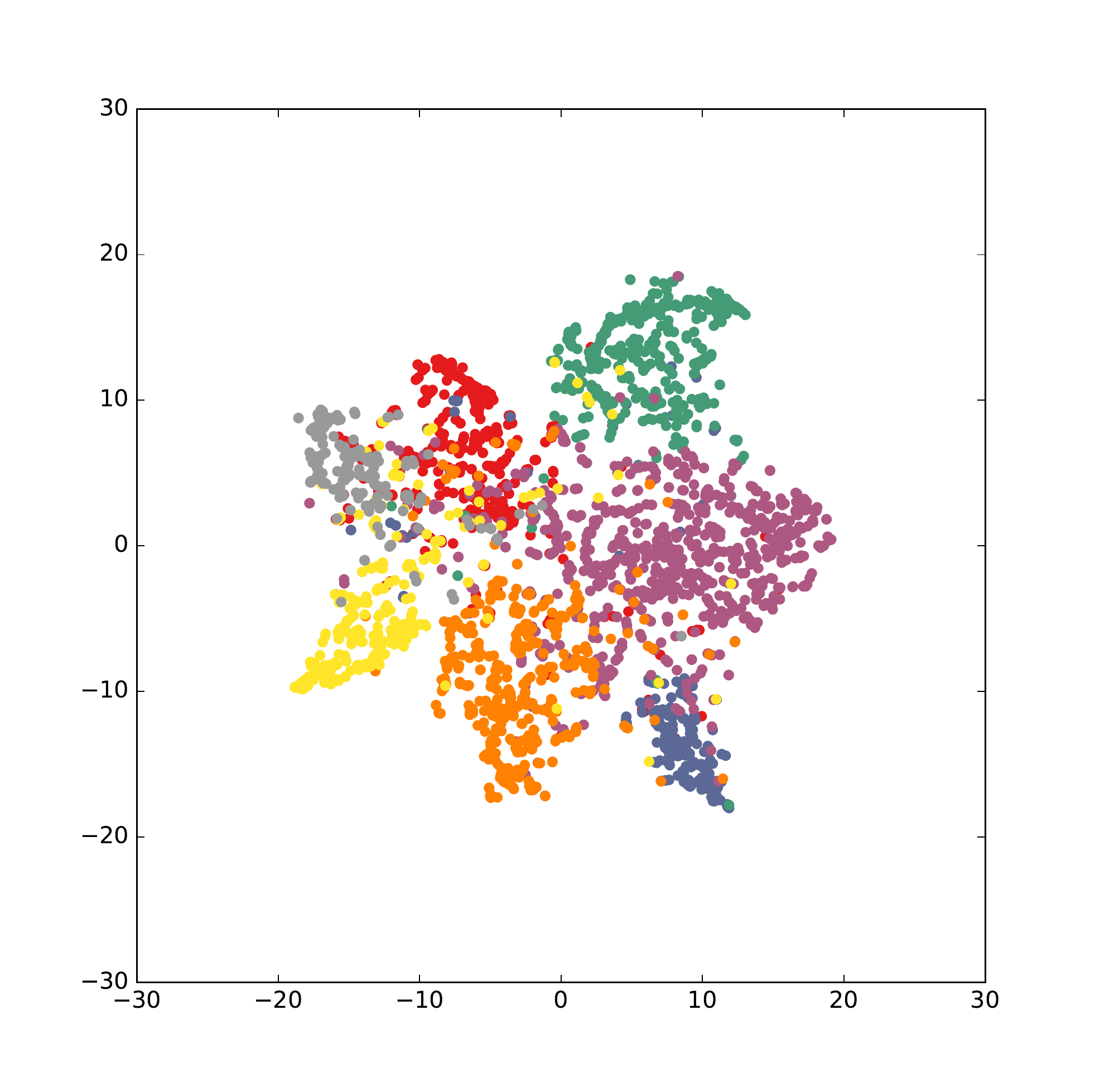}
        \caption{Hidden layer activations}
        \label{fig:model-b}
    \end{subfigure}

    \caption{\emph{Left}: Schematic depiction of multi-layer Graph Convolutional Network (GCN) for semi-supervised learning with $C$ input channels and $F$ feature maps in the output layer. The graph structure (edges shown as black lines) is shared over layers, labels are denoted by $Y_i$. \emph{Right}: t-SNE \citep{maaten2008visualizing} visualization of hidden layer activations of a two-layer GCN trained on the Cora dataset \citep{sen2008collective} using $5\%$ of labels. Colors denote document class.}
    \label{fig:model}
\end{figure}

Having introduced a simple, yet flexible model $f(X,A)$ for efficient information propagation on graphs, we can return to the problem of semi-supervised node classification. As outlined in the introduction, we can relax certain assumptions typically made in graph-based semi-supervised learning by conditioning our model $f(X,A)$ both on the data $X$ and on the adjacency matrix $A$ of the underlying graph structure. We expect this setting to be especially powerful in scenarios where the adjacency matrix contains information not present in the data $X$, such as citation links between documents in a citation network or relations in a knowledge graph. The overall model, a multi-layer GCN for semi-supervised learning, is schematically depicted in Figure \ref{fig:model}.

\subsection{Example} \label{sec:model-example} In the following, we consider a two-layer GCN for semi-supervised node classification on a graph with a symmetric adjacency matrix $A$ (binary or weighted). We first calculate $\hat{A} = \tilde{D}^{-\frac{1}{2}}\tilde{A}\tilde{D}^{-\frac{1}{2}}$ in a pre-processing step. Our forward model then takes the simple form:
\begin{equation}
Z= f(X,A) =  \mathrm{softmax}\!\left(\hat{A} \,\, \mathrm{ReLU}\!\left(\hat{A} X W^{(0)} \right) W^{(1)} \right) \, .
\label{eq:two-layer-gcn}
\end{equation}
Here, $W^{(0)}\in \mathbb{R}^{C\times H}$ is an input-to-hidden weight matrix for a hidden layer with $H$ feature maps. $W^{(1)}\in \mathbb{R}^{H\times F}$ is a hidden-to-output weight matrix. The softmax activation function, defined as $\mathrm{softmax}(x_i)=\frac{1}{\mathcal{Z}}\exp(x_i)$ with $\mathcal{Z}=\sum_i \exp(x_i)$, is applied row-wise. For semi-supervised multi-class classification, we then evaluate the cross-entropy error over all labeled examples:
\begin{equation}
\mathcal{L}= -\sum_{l\in\mathcal{Y}_L}\sum_{f=1}^F Y_{lf} \ln Z_{lf}  \, ,
\label{eq:}
\end{equation}
where $\mathcal{Y}_L$ is the set of node indices that have labels.

The neural network weights $W^{(0)}$ and $W^{(1)}$ are trained using gradient descent. In this work, we perform batch gradient descent using the full dataset for every training iteration, which is a viable option as long as datasets fit in memory. Using a sparse representation for $A$, memory requirement is $\mathcal{O}(|\mathcal{E}|)$, \ie linear in the number of edges. Stochasticity in the training process is introduced via dropout \citep{srivastava2014dropout}. We leave memory-efficient extensions with mini-batch stochastic gradient descent for future work.

\subsection{Implementation} In practice, we make use of TensorFlow \citep{tensorflow2015-whitepaper} for an efficient GPU-based implementation\footnote{Code to reproduce our experiments is available at \url{https://github.com/tkipf/gcn}.} of \eq \ref{eq:two-layer-gcn} using sparse-dense matrix multiplications. The computational complexity of evaluating \eq \ref{eq:two-layer-gcn} is then $\mathcal{O}(|\mathcal{E}|CHF)$, \ie linear in the number of graph edges.

\section{Related Work}
Our model draws inspiration both from the field of graph-based semi-supervised learning and from recent work on neural networks that operate on graphs. In what follows, we provide a brief overview on related work in both fields.

\subsection{Graph-Based Semi-Supervised Learning}
A large number of approaches for semi-supervised learning using graph representations have been proposed in recent years, most of which fall into two broad categories: methods that use some form of explicit graph Laplacian regularization and graph embedding-based approaches. Prominent examples for graph Laplacian regularization include label propagation \citep{zhu2003semi}, manifold regularization \citep{belkin2006manifold} and deep semi-supervised embedding \citep{weston2012deep}. 

Recently, attention has shifted to models that learn graph embeddings with methods inspired by the skip-gram model \citep{mikolov2013distributed}. DeepWalk \citep{perozzi2014deepwalk} learns embeddings via the prediction of the local neighborhood of nodes, sampled from random walks on the graph. LINE \citep{tang2015line} and node2vec \citep{grovernode2vec} extend DeepWalk with more sophisticated random walk or breadth-first search schemes. For all these methods, however, a multi-step pipeline including random walk generation and semi-supervised training is required where each step has to be optimized separately. Planetoid \citep{yang2016revisiting} alleviates this by injecting label information in the process of learning embeddings.

\subsection{Neural Networks on Graphs} Neural networks that operate on graphs have previously been introduced in \cite{gori2005new, scarselli2009graph} as a form of recurrent neural network. Their framework requires the repeated application of contraction maps as propagation functions until node representations reach a stable fixed point. This restriction was later alleviated in \cite{li2015gated} by introducing modern practices for recurrent neural network training to the original graph neural network framework. \cite{duvenaud2015convolutional} introduced a convolution-like propagation rule on graphs and methods for graph-level classification. Their approach requires to learn node degree-specific weight matrices which does not scale to large graphs with wide node degree distributions. Our model instead uses a single weight matrix per layer and deals with varying node degrees through an appropriate normalization of the adjacency matrix (see Section \ref{sec:model-example}).

A related approach to node classification with a graph-based neural network was recently introduced in \cite{atwood2016diffusion}. They report $\mathcal{O}(N^2)$ complexity, limiting the range of possible applications. In a different yet related model, \cite{niepert2016learning} convert graphs locally into sequences that are fed into a conventional 1D convolutional neural network, which requires the definition of a node ordering in a pre-processing step.

Our method is based on spectral graph convolutional neural networks, introduced in \cite{bruna2014spectral} and later extended by \cite{defferrard2016convolutional} with fast localized convolutions. In contrast to these works, we consider here the task of transductive node classification within networks of significantly larger scale. We show that in this setting, a number of simplifications (see Section \ref{sec:linear-model}) can be introduced to the original frameworks of \cite{bruna2014spectral} and \cite{defferrard2016convolutional} that improve scalability and classification performance in large-scale networks.

\section{Experiments}
We test our model in a number of experiments: semi-supervised document classification in citation networks, semi-supervised entity classification in a bipartite graph extracted from a knowledge graph, an evaluation of various graph propagation models and a run-time analysis on random graphs.

\subsection{Datasets}
\label{sec:datasets}
We closely follow the experimental setup in \cite{yang2016revisiting}. Dataset statistics are summarized in Table \ref{tab:datasets}. In the citation network datasets---Citeseer, Cora and Pubmed \citep{sen2008collective}---nodes are documents and edges are citation links. Label rate denotes the number of labeled nodes that are used for training divided by the total number of nodes in each dataset. NELL \citep{carlson2010toward, yang2016revisiting} is a bipartite graph dataset extracted from a knowledge graph with 55,864 relation nodes and 9,891 entity nodes.

\begin{table}[htp]
\centering
\caption{\label{tab:datasets}Dataset statistics, as reported in \cite{yang2016revisiting}.}
\begin{tabular}{l c r r r r r}
\textbf{Dataset} & \textbf{Type} & \textbf{Nodes} & \textbf{Edges} & \textbf{Classes} & \textbf{Features} & \textbf{Label rate}  \\[0.05em]\hline \\[-0.8em]
Citeseer & Citation network & 3,327 & 4,732 & $6$ & 3,703 & $0.036$ \\
Cora & Citation network & 2,708 & 5,429 & $7$ & 1,433 & $0.052$ \\
Pubmed & Citation network& 19,717 & 44,338 & $3$ & 500 & $0.003$ \\
NELL & Knowledge graph & 65,755 & 266,144 & $210$ & 5,414 & $0.001$ \\
\end{tabular}
\end{table}

\paragraph{Citation networks} We consider three citation network datasets: Citeseer, Cora and Pubmed \citep{sen2008collective}. The datasets contain sparse bag-of-words feature vectors for each document and a list of citation links between documents. We treat the citation links as (undirected) edges and construct a binary, symmetric adjacency matrix $A$. Each document has a class label. For training, we only use 20 labels per class, but all feature vectors.

\paragraph{NELL} NELL is a dataset extracted from the knowledge graph introduced in \citep{carlson2010toward}. A knowledge graph is a set of entities connected with directed, labeled edges (relations). We follow the pre-processing scheme as described in \cite{yang2016revisiting}. We assign separate relation nodes $r_1$ and $r_2$ for each entity pair $(e_1, r, e_2)$ as $(e_1, r_1)$ and $(e_2, r_2)$. Entity nodes are described by sparse feature vectors. We extend the number of features in NELL by assigning a unique one-hot representation for every relation node, effectively resulting in a 61,278-dim sparse feature vector per node. The semi-supervised task here considers the extreme case of only a single labeled example per class in the training set. We construct a binary, symmetric adjacency matrix from this graph by setting entries $A_{ij}=1$, if one or more edges are present between nodes $i$ and $j$.

\paragraph{Random graphs} We simulate random graph datasets of various sizes for experiments where we measure training time per epoch. For a dataset with $N$ nodes we create a random graph assigning $2N$ edges uniformly at random. We take the identity matrix $I_N$ as input feature matrix $X$, thereby implicitly taking a featureless approach where the model is only informed about the identity of each node, specified by a unique one-hot vector. We add dummy labels $Y_i = 1$ for every node.

\subsection{Experimental Set-Up}
Unless otherwise noted, we train a two-layer GCN as described in Section \ref{sec:model-example} and evaluate prediction accuracy on a test set of 1,000 labeled examples. We provide additional experiments using deeper models with up to 10 layers in Appendix \ref{sec:depth}. We choose the same dataset splits as in \cite{yang2016revisiting} with an additional validation set of 500 labeled examples for hyperparameter optimization (dropout rate for all layers, L2 regularization factor for the first GCN layer and number of hidden units). We do not use the validation set labels for training.

For the citation network datasets, we optimize hyperparameters on Cora only and use the same set of parameters for Citeseer and Pubmed. We train all models for a maximum of 200 epochs (training iterations) using Adam \citep{kingma2014adam} with a learning rate of $0.01$ and early stopping with a window size of $10$, \ie we stop training if the validation loss does not decrease for 10 consecutive epochs. We initialize weights using the initialization described in \cite{glorot2010understanding} and accordingly (row-)normalize input feature vectors. On the random graph datasets, we use a hidden layer size of 32 units and omit regularization (\ie neither dropout nor L2 regularization).

\subsection{Baselines}

We compare against the same baseline methods as in \cite{yang2016revisiting}, \ie label propagation (LP) \citep{zhu2003semi}, semi-supervised embedding (SemiEmb) \citep{weston2012deep}, manifold regularization (ManiReg) \citep{belkin2006manifold} and skip-gram based graph embeddings (DeepWalk) \citep{perozzi2014deepwalk}. We omit TSVM \citep{joachims1999transductive}, as it does not scale to the large number of classes in one of our datasets.

We further compare against the iterative classification algorithm (ICA) proposed in \cite{lu2003link} in conjunction with two logistic regression classifiers, one for local node features alone and one for relational classification using local features and an aggregation operator as described in \cite{sen2008collective}. We first train the local classifier using all labeled training set nodes and use it to bootstrap class labels of unlabeled nodes for relational classifier training. We run iterative classification (relational classifier) with a random node ordering for 10 iterations on all unlabeled nodes (bootstrapped using the local classifier). L2 regularization parameter and aggregation operator (\textit{count} \vs \textit{prop}, see \cite{sen2008collective}) are chosen based on validation set performance for each dataset separately.

Lastly, we compare against Planetoid \citep{yang2016revisiting}, where we always choose their best-performing model variant (transductive \vs inductive) as a baseline.

\section{Results}
\label{sec:results}

\subsection{Semi-Supervised Node Classification}
Results are summarized in Table \ref{tab:results_node}. Reported numbers denote classification accuracy in percent. For ICA, we report the mean accuracy of 100 runs with random node orderings. Results for all other baseline methods are taken from the Planetoid paper \citep{yang2016revisiting}. Planetoid* denotes the best model for the respective dataset out of the variants presented in their paper.

\begin{table}[htp]
\centering
\caption{\label{tab:results_node}Summary of results in terms of classification accuracy (in percent).}
\begin{tabular}{l l l l l}
\textbf{Method} & \textbf{Citeseer} & \textbf{Cora} & \textbf{Pubmed} & \textbf{NELL} \\[0.05em]\hline \\[-0.8em]
ManiReg [\citenum{belkin2006manifold}] & $60.1$ & $59.5$ & $70.7$ & $21.8$ \\
SemiEmb [\citenum{weston2012deep}] & $59.6$ & $59.0$ & $71.1$ & $26.7$ \\
LP [\citenum{zhu2003semi}] & $45.3$ & $68.0$ & $63.0$ & $26.5$ \\
DeepWalk [\citenum{perozzi2014deepwalk}] & $43.2$ & $67.2$ & $65.3$ & $58.1$ \\
ICA [\citenum{lu2003link}] & $69.1$ & $75.1$ & $73.9$ & $23.1$ \\
Planetoid* [\citenum{yang2016revisiting}] & $64.7$ (26s) & $75.7$ (13s) & $77.2$ (25s) & $61.9$ (185s) \\
\textbf{GCN} (this paper) & $\mathbf{70.3}$ (7s) & $\mathbf{81.5}$ (4s) & $\mathbf{79.0}$ (38s) & $\mathbf{66.0}$ (48s) \\[0.05em]\hline \\[-0.8em]
GCN (rand.\@\xspace splits) & $67.9 \pm 0.5$ & $80.1 \pm 0.5$ & $78.9 \pm 0.7$ & $58.4 \pm 1.7$ \\
\end{tabular}
\end{table}

We further report wall-clock training time in seconds until convergence (in brackets) for our method (incl.~evaluation of validation error) and for Planetoid. For the latter, we used an implementation provided by the authors\footnote{\url{https://github.com/kimiyoung/planetoid}} and trained on the same hardware (with GPU) as our GCN model. We trained and tested our model on the same dataset splits as in \cite{yang2016revisiting} and report mean accuracy of 100 runs with random weight initializations. We used the following sets of hyperparameters for Citeseer, Cora and Pubmed: 0.5 (dropout rate), $5\cdot 10^{-4}$ (L2 regularization) and $16$ (number of hidden units); and for NELL: 0.1 (dropout rate), $1\cdot 10^{-5}$ (L2 regularization) and $64$ (number of hidden units).

In addition, we report performance of our model on 10 randomly drawn dataset splits of the same size as in \cite{yang2016revisiting}, denoted by GCN (rand.\@\xspace splits). Here, we report mean and standard error of prediction accuracy on the test set split in percent.

\subsection{Evaluation of Propagation Model}
We compare different variants of our proposed per-layer propagation model on the citation network datasets. We follow the experimental set-up described in the previous section. Results are summarized in Table \ref{tab:results_model_eval}. The propagation model of our original GCN model is denoted by \textit{renormalization trick} (in bold). In all other cases, the propagation model of both neural network layers is replaced with the model specified under \textit{propagation model}. Reported numbers denote mean classification accuracy for 100 repeated runs with random weight matrix initializations. In case of multiple variables $\Theta_i$ per layer, we impose L2 regularization on all weight matrices of the first layer.

\begin{table}[htp]
\centering
\caption{\label{tab:results_model_eval}Comparison of propagation models.}
\begin{tabular}{l c c r r r}
\textbf{Description} & & \textbf{Propagation model} & \textbf{Citeseer} & \textbf{Cora} & \textbf{Pubmed} \\[0.05em]\hline \\[-0.8em]

\multirow{2}{*}{Chebyshev filter (\eq \ref{eq:fourier-conv-approx})} & $K=3$  & \multirow{2}{*}{$\sum_{k=0}^K  T_k(\tilde{L}) X \Theta_k$} & $69.8$& $79.5$ & $74.4$    \\
  & $K=2$ &  & $69.6$& $81.2$ & $73.8$    \\[0.05em]\hline \\[-0.9em]
\multicolumn{2}{l}{$1^{\text{st}}$-order model (\eq \ref{eq:fourier-conv-approx2})}  & $X\Theta_0 + D^{-\frac{1}{2}}AD^{-\frac{1}{2}}X\Theta_1$ & $68.3$& $80.0$ & $77.5$ \\
\multicolumn{2}{l}{Single parameter (\eq \ref{eq:fourier-conv-approx3})} & $(I_N + D^{-\frac{1}{2}}AD^{-\frac{1}{2}})X\Theta$ & $69.3$& $79.2$ & $77.4$ \\
\multicolumn{2}{l}{\textbf{Renormalization trick} (\eq \ref{eq:fourier-conv-approx4})}& $\tilde{D}^{-\frac{1}{2}}\tilde{A}\tilde{D}^{-\frac{1}{2}}X\Theta$ & $\mathbf{70.3}$ & $\mathbf{81.5}$ & $\mathbf{79.0}$   \\[0.25em]\hline \\[-0.9em]
\multicolumn{2}{l}{$1^{\text{st}}$-order term only}
 & $D^{-\frac{1}{2}}AD^{-\frac{1}{2}}X\Theta$ & $68.7$ & $80.5$ & $77.8$ \\
\multicolumn{2}{l}{Multi-layer perceptron} & $X\Theta$ & $46.5$ & $55.1$ & $71.4$ \\
\end{tabular}
\end{table}

\newpage
\subsection{Training Time per Epoch}
\begin{wrapfigure}{r}{0.5\textwidth}
\vspace{-3em}
    \centering
    \includegraphics[scale=0.56, trim={0.45cm 0.45cm 0.45cm 0}, clip]{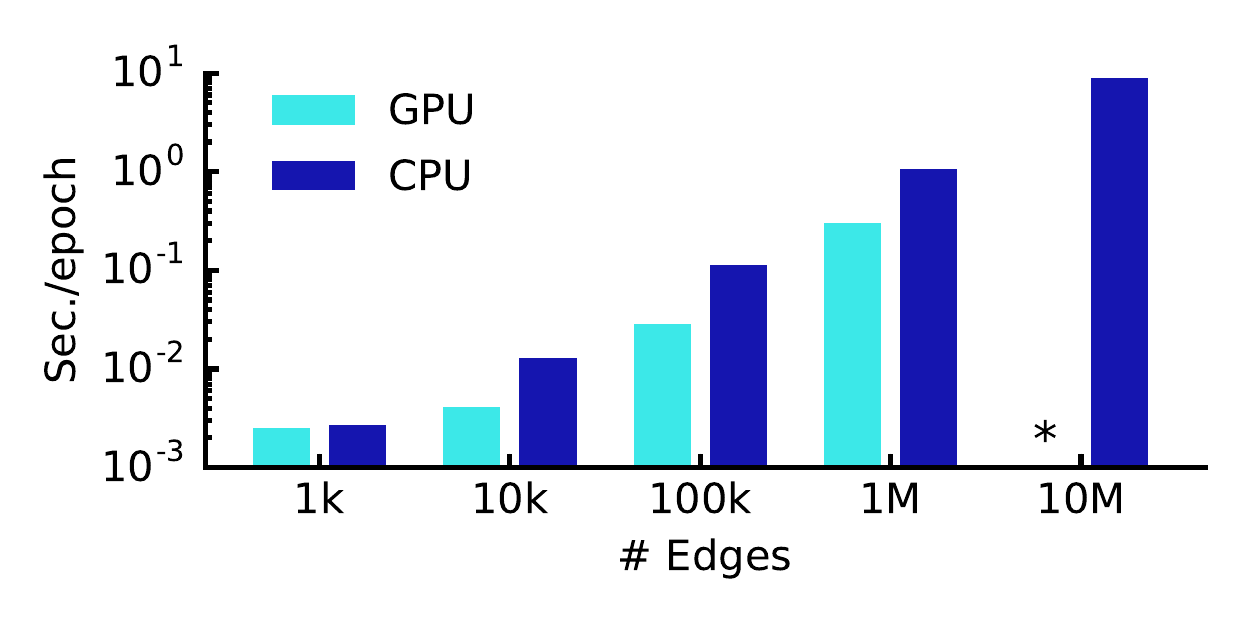}
    \caption{Wall-clock time per epoch for random graphs. (*) indicates out-of-memory error.}
    \label{fig:epoch_timing}
    \vspace{-1em}
\end{wrapfigure}

Here, we report results for the mean training time per epoch (forward pass, cross-entropy calculation, backward pass) for 100 epochs on simulated random graphs, measured in seconds wall-clock time. See Section \ref{sec:datasets} for a detailed description of the random graph dataset used in these experiments. We compare results on a GPU and on a CPU-only implementation\footnote{Hardware used: 16-core Intel\registered Xeon\registered CPU E5-2640 v3 @ 2.60GHz, GeForce\registered GTX TITAN X} in TensorFlow \citep{tensorflow2015-whitepaper}. Figure \ref{fig:epoch_timing} summarizes the results.

\section{Discussion}

\subsection{Semi-Supervised Model}

In the experiments demonstrated here, our method for semi-supervised node classification outperforms recent related methods by a significant margin. Methods based on graph-Laplacian regularization \citep{zhu2003semi, belkin2006manifold, weston2012deep} are most likely limited due to their assumption that edges encode mere similarity of nodes. Skip-gram based methods on the other hand are limited by the fact that they are based on a multi-step pipeline which is difficult to optimize. Our proposed model can overcome both limitations, while still comparing favorably in terms of efficiency (measured in wall-clock time) to related methods. Propagation of feature information from neighboring nodes in every layer improves classification performance in comparison to methods like ICA \citep{lu2003link}, where only label information is aggregated.


We have further demonstrated that the proposed renormalized propagation model (\eq \ref{eq:fourier-conv-approx4}) offers both improved efficiency (fewer parameters and operations, such as multiplication or addition) and better predictive performance on a number of datasets compared to a na\"ive $1^{\text{st}}$-order model (\eq \ref{eq:fourier-conv-approx2}) or higher-order graph convolutional models using Chebyshev polynomials (\eq \ref{eq:fourier-conv-approx}).

\subsection{Limitations and Future Work}
Here, we describe several limitations of our current model and outline how these might be overcome in future work.

\paragraph{Memory requirement}
In the current setup with full-batch gradient descent, memory requirement grows linearly in the size of the dataset. We have shown that for large graphs that do not fit in GPU memory, training on CPU can still be a viable option. Mini-batch stochastic gradient descent can alleviate this issue. The procedure of generating mini-batches, however, should take into account the number of layers in the GCN model, as the $K^{\text{th}}$-order neighborhood for a GCN with $K$ layers has to be stored in memory for an exact procedure. For very large and densely connected graph datasets, further approximations might be necessary.

\paragraph{Directed edges and edge features}
Our framework currently does not naturally support edge features and is limited to undirected graphs (weighted or unweighted). Results on NELL however show that it is possible to handle both directed edges and edge features by representing the original directed graph as an undirected bipartite graph with additional nodes that represent edges in the original graph (see Section \ref{sec:datasets} for details).

\paragraph{Limiting assumptions}
Through the approximations introduced in Section \ref{sec:fast-convs}, we implicitly assume locality (dependence on the $K^{\text{th}}$-order neighborhood for a GCN with $K$ layers) and equal importance of self-connections \vs edges to neighboring nodes. For some datasets, however, it might be beneficial to introduce a trade-off parameter $\lambda$ in the definition of $\tilde{A}$:
\begin{equation}
  \tilde{A} = A + \lambda I_N\, .
\label{eq:trade-off-param}
\end{equation}
This parameter now plays a similar role as the trade-off parameter between supervised and unsupervised loss in the typical semi-supervised setting (see \eq \ref{eq:graph-reg}). Here, however, it can be learned via gradient descent.

\section{Conclusion}
We have introduced a novel approach for semi-supervised classification on graph-structured data. Our GCN model uses an efficient layer-wise propagation rule that is based on a first-order approximation of spectral convolutions on graphs. Experiments on a number of network datasets suggest that the proposed GCN model is capable of encoding both graph structure and node features in a way useful for semi-supervised classification. In this setting, our model outperforms several recently proposed methods by a significant margin, while being computationally efficient.

\subsubsection*{Acknowledgments}

We would like to thank Christos Louizos, Taco Cohen, Joan Bruna, Zhilin Yang, Dave Herman, Pramod Sinha and Abdul-Saboor Sheikh for helpful discussions. This research was funded by SAP.

\bibliography{references}
\bibliographystyle{iclr2017_conference}

\appendix

\section{Relation to Weisfeiler-Lehman Algorithm}
\label{sec:wl}
A neural network model for graph-structured data should ideally be able to learn representations of nodes in a graph, taking both the graph structure and feature description of nodes into account. A well-studied framework for the unique assignment of node labels given a graph and (optionally) discrete initial node labels is provided by the 1-dim Weisfeiler-Lehman (WL-1) algorithm  \citep{weisfeiler1968reduction}:

\begin{algorithm}[H]
\KwIn{Initial node coloring $(h^{(0)}_1, h^{(0)}_2, ..., h^{(0)}_N )$}
\KwOut{Final node coloring $(h^{(T)}_1, h^{(T)}_2, ..., h^{(T)}_N )$}
t $\leftarrow$ 0\;
\Repeat{stable node coloring is reached}{
          \For{$v_i \in \mathcal{V}$} {
			$h^{(t+1)}_i \leftarrow \mathrm{hash}\left(\sum_{j\in\mathcal{N}_i} h^{(t)}_j\right) $\;
          }
    $t \leftarrow t+1$\;
    }
    \caption{{\bf WL-1 algorithm \citep{weisfeiler1968reduction}} \label{alg:wl1}}
\end{algorithm}

Here, $h_i^{(t)}$ denotes the coloring (label assignment) of node $v_i$ (at iteration $t$) and $\mathcal{N}_i$ is its set of neighboring node indices (irrespective of whether the graph includes self-connections for every node or not). $\mathrm{hash}(\cdot)$ is a hash function. For an in-depth mathematical discussion of the WL-1 algorithm see, \eg, \cite{douglas2011weisfeiler}.

We can replace the hash function in Algorithm \ref{alg:wl1} with a neural network layer-like differentiable function with trainable parameters as follows:
\begin{equation}
  h^{(l+1)}_i = \sigma \left( \sum_{j\in\mathcal{N}_i} \frac{1}{c_{ij}}h^{(l)}_jW^{(l)} \right) \, ,
\label{eq:diff-model}
\end{equation}
where $c_{ij}$ is an appropriately chosen normalization constant for the edge $(v_i,v_j)$. Further, we can take $h^{(l)}_i$ now to be a vector of \emph{activations} of node $i$ in the $l^{\text{th}}$ neural network layer. $W^{(l)}$ is a layer-specific weight matrix and $\sigma(\cdot)$ denotes a differentiable, non-linear activation function.

By choosing $c_{ij}=\sqrt{d_i d_j}$, where $d_i=|\mathcal{N}_i|$ denotes the degree of node $v_i$, we recover the propagation rule of our Graph Convolutional Network (GCN) model in vector form (see \eq \ref{eq:gcn-layer})\footnote{Note that we here implicitly assume that self-connections have already been added to every node in the graph (for a clutter-free notation).}.

This---loosely speaking---allows us to interpret our GCN model as a differentiable and parameterized generalization of the 1-dim Weisfeiler-Lehman algorithm on graphs.

\subsection{Node Embeddings with Random Weights}
From the analogy with the Weisfeiler-Lehman algorithm, we can understand that even an untrained GCN model with random weights can serve as a powerful feature extractor for nodes in a graph. As an example, consider the following 3-layer GCN model:
\begin{equation}
Z= \tanh\!\left(\hat{A} \, \tanh\!\left(\hat{A}\,\tanh\!\left(\hat{A} X W^{(0)}\right) W^{(1)} \right) W^{(2)} \right) \, ,
\label{eq:three-layer-gcn}
\end{equation}
with weight matrices $W^{(l)}$ initialized at random using the initialization described in \cite{glorot2010understanding}. $\hat{A}$, $X$ and $Z$ are defined as in Section \ref{sec:model-example}.

\begin{figure}[htbp]
\centering
\begin{subfigure}[b]{0.5\textwidth}
    \centering
    \includegraphics[width=0.95\textwidth, trim={5cm 3.2cm 4.9cm 4.4cm}, clip]{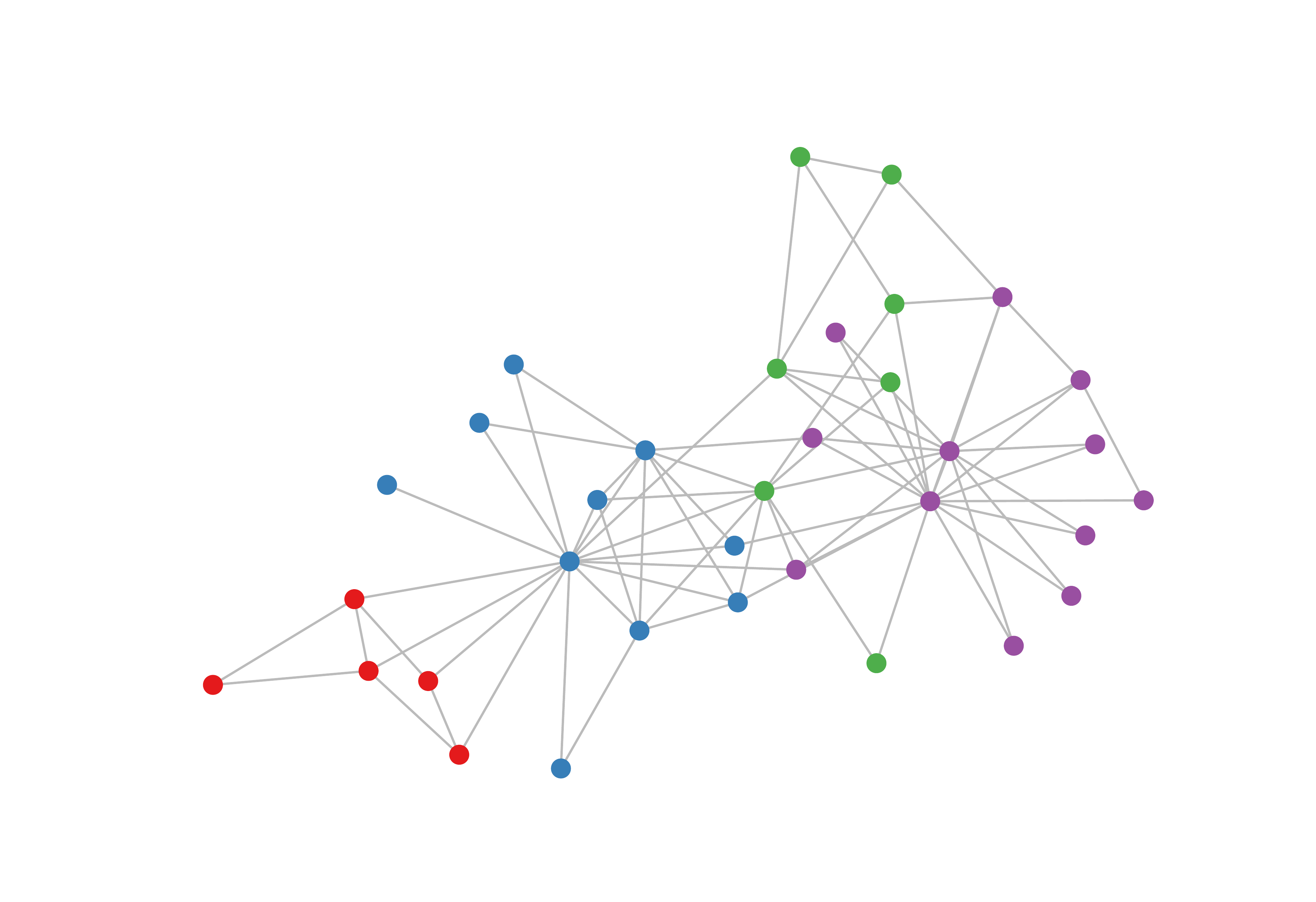}
    \caption{Karate club network}
    \label{fig:karate-club-a}
\end{subfigure}%
~
\begin{subfigure}[b]{0.5\textwidth}
    \centering
    \includegraphics[width=\textwidth, trim={0 1cm 0 0}, clip]{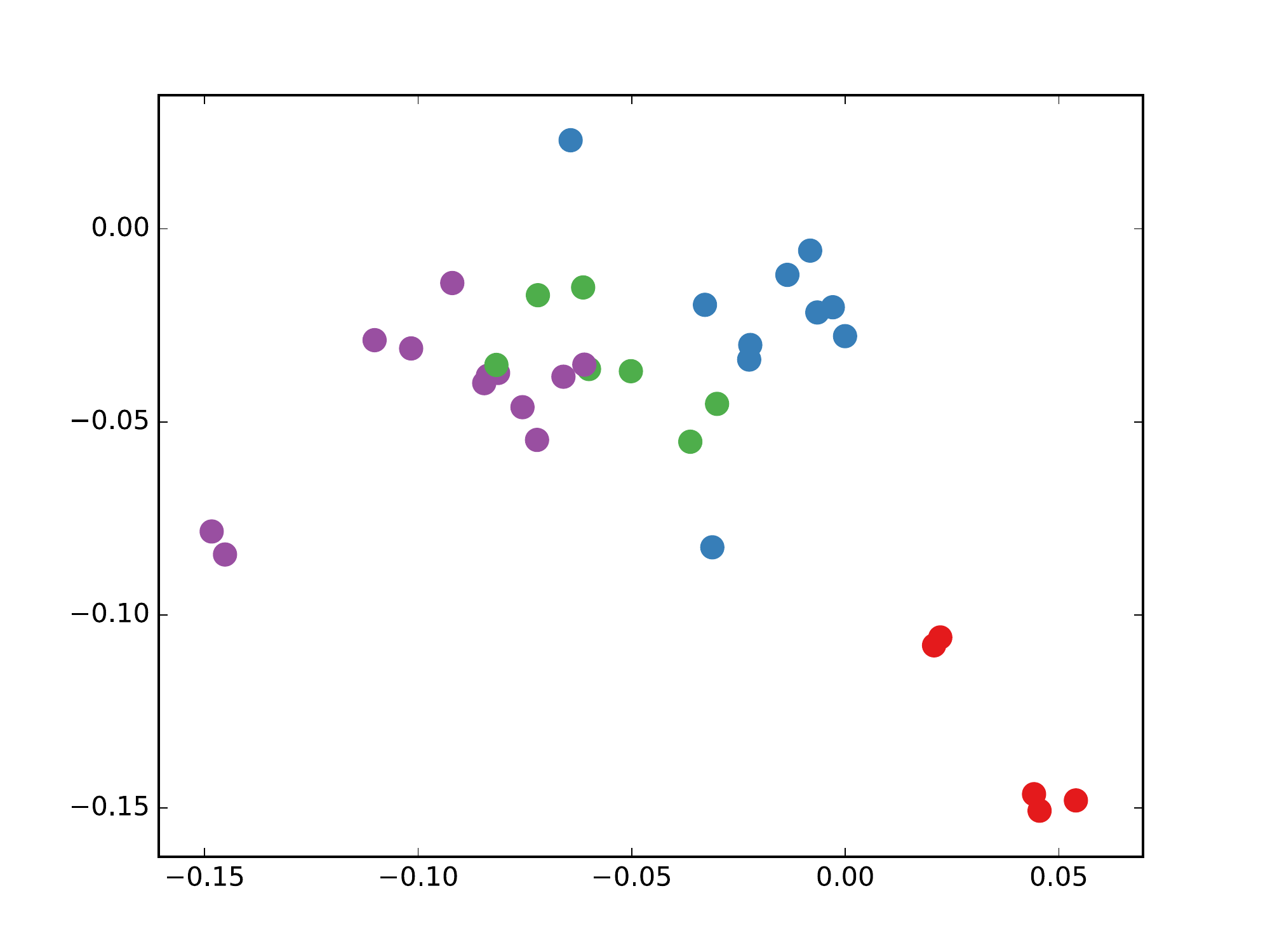}
    \caption{Random weight embedding}
    \label{fig:karate-club-b}
\end{subfigure}
\caption{\textit{Left}: Zachary's karate club network \citep{zachary1977information}, colors denote communities obtained via modularity-based clustering \citep{brandes2008modularity}. \textit{Right}: Embeddings obtained from an untrained 3-layer GCN model (\eq \ref{eq:three-layer-gcn}) with random weights applied to the karate club network. Best viewed on a computer screen.}
\label{fig:karate-club}
\end{figure}

We apply this model on Zachary's karate club network \citep{zachary1977information}. This graph contains 34 nodes, connected by 154 (undirected and unweighted) edges. Every node is labeled by one of four classes, obtained via modularity-based clustering \citep{brandes2008modularity}. See Figure \ref{fig:karate-club-a} for an illustration.

We take a featureless approach by setting $X=I_N$, where $I_N$ is the $N$ by $N$ identity matrix. $N$ is the number of nodes in the graph. Note that nodes are randomly ordered (\ie ordering contains no information). Furthermore, we choose a hidden layer dimensionality\footnote{We originally experimented with a hidden layer dimensionality of $2$ (\ie same as output layer), but observed that a dimensionality of $4$ resulted in less frequent saturation of $\tanh(\cdot)$ units and therefore visually more pleasing results.} of $4$ and a two-dimensional output (so that the output can immediately be visualized in a 2-dim plot).

Figure \ref{fig:karate-club-b} shows a representative example of node embeddings (outputs $Z$) obtained from an untrained GCN model applied to the karate club network. These results are comparable to embeddings obtained from DeepWalk \citep{perozzi2014deepwalk}, which uses a more expensive unsupervised training procedure.

\subsection{Semi-Supervised Node Embeddings}
\begin{figure}[htbp]
\centering
\begin{subfigure}[b]{0.5\textwidth}
    \centering
    \includegraphics[width=\textwidth, trim={0 1.5cm 0 0}, clip]{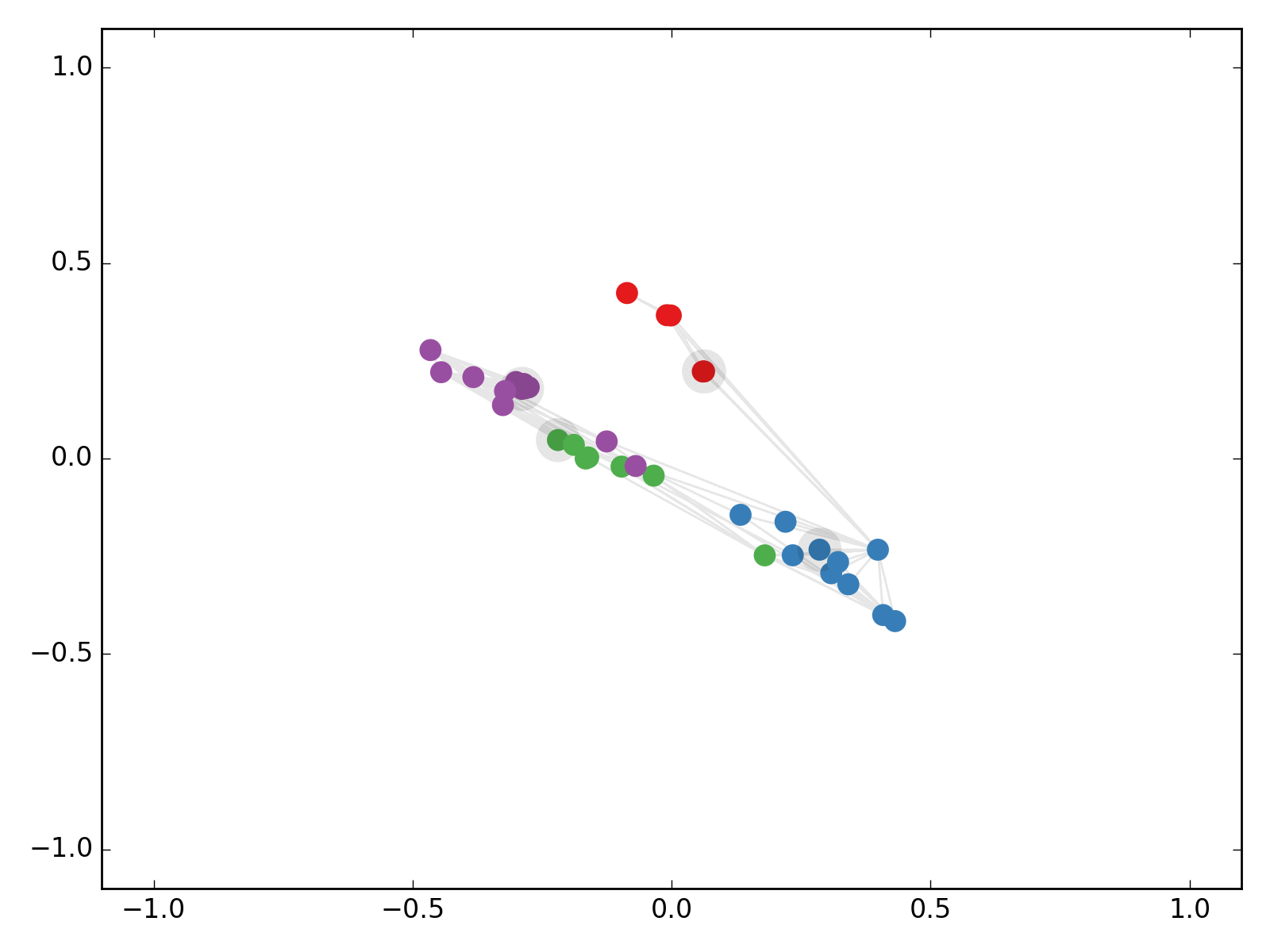}
    \caption{Iteration 25}
    \label{fig:semi-emb-a}
\end{subfigure}%
~
\begin{subfigure}[b]{0.5\textwidth}
    \centering
    \includegraphics[width=\textwidth, trim={0 1.5cm 0 0}, clip]{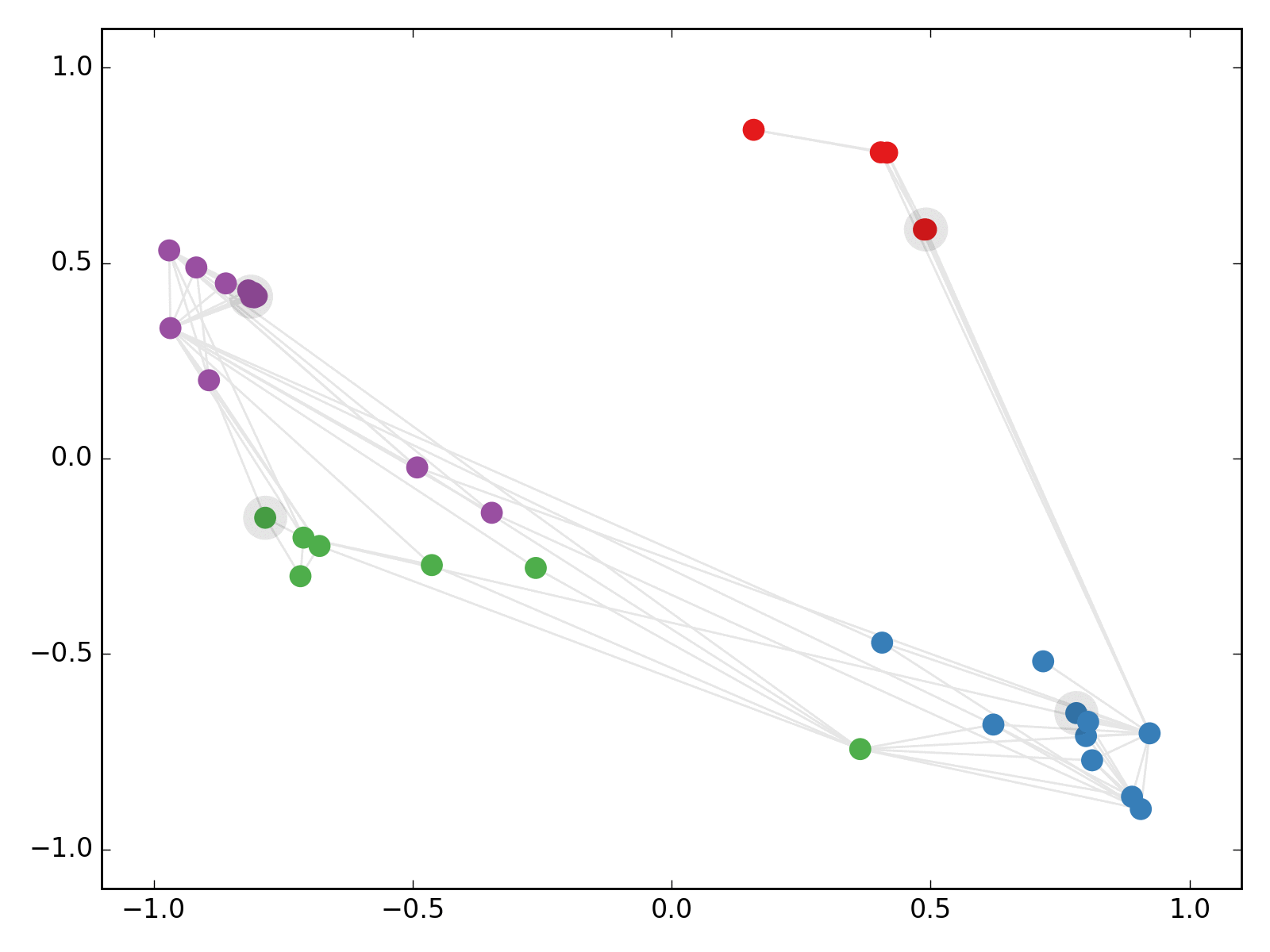}
    \caption{Iteration 50}
    \label{fig:semi-emb-b}
\end{subfigure}%
\vspace{0.5em}
\begin{subfigure}[b]{0.5\textwidth}
    \centering
    \includegraphics[width=\textwidth, trim={0 1.5cm 0 0}, clip]{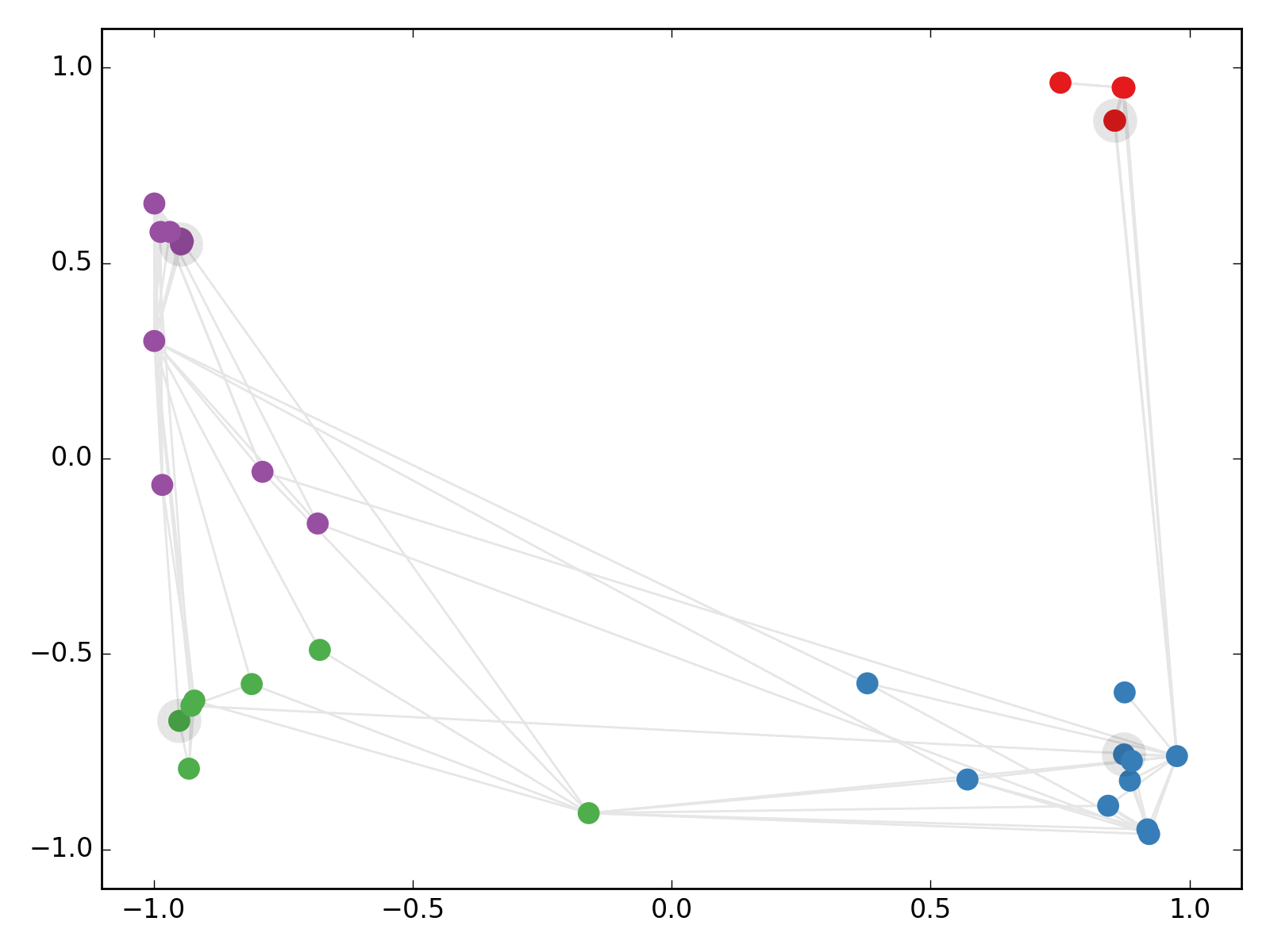}
    \caption{Iteration 75}
    \label{fig:semi-emb-c}
\end{subfigure}%
~
\begin{subfigure}[b]{0.5\textwidth}
    \centering
    \includegraphics[width=\textwidth, trim={0 1.5cm 0 0}, clip]{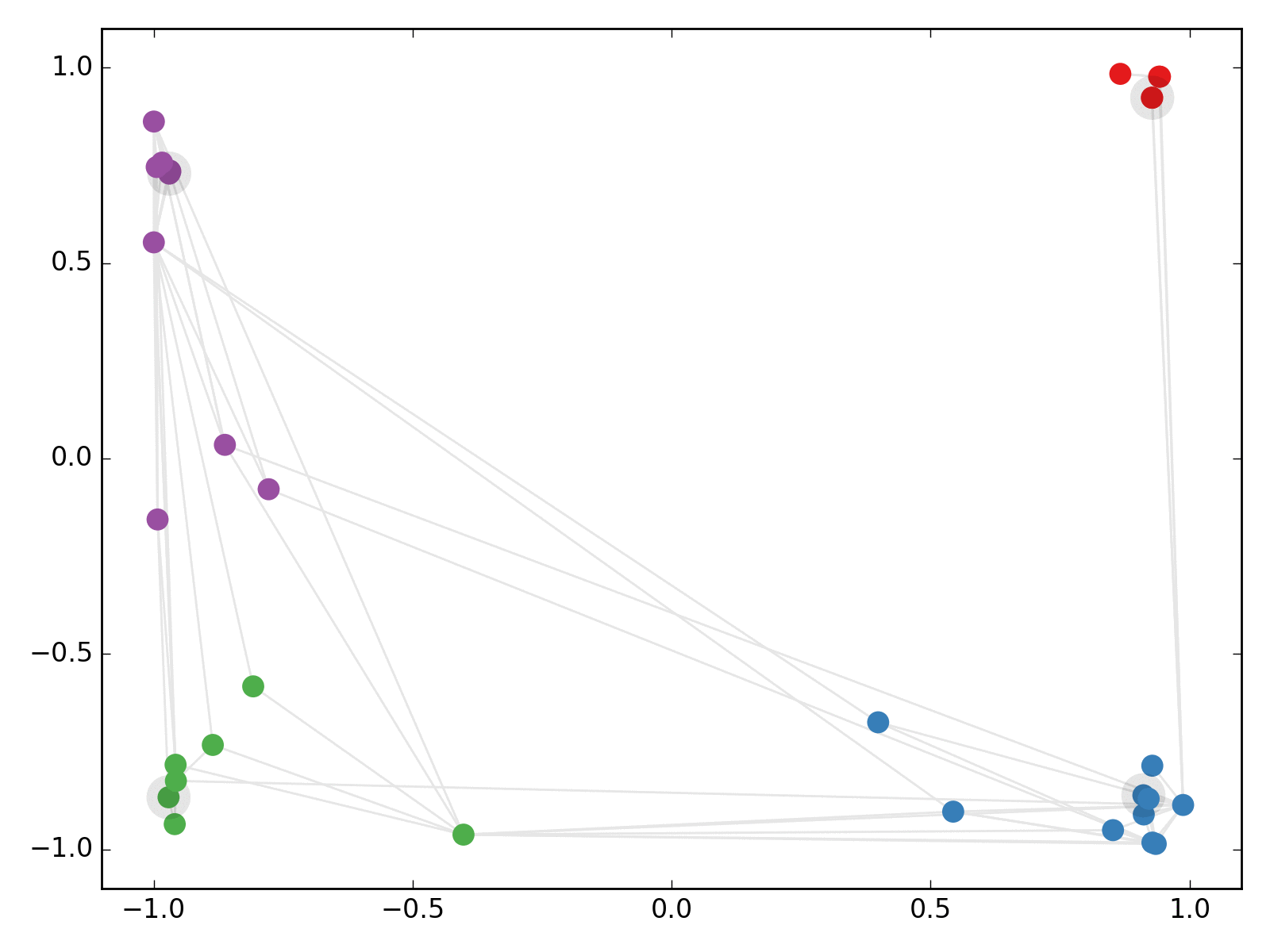}
    \caption{Iteration 100}
    \label{fig:semi-emb-d}
\end{subfigure}%
\vspace{0.5em}
\begin{subfigure}[b]{0.5\textwidth}
    \centering
    \includegraphics[width=\textwidth, trim={0 1.5cm 0 0}, clip]{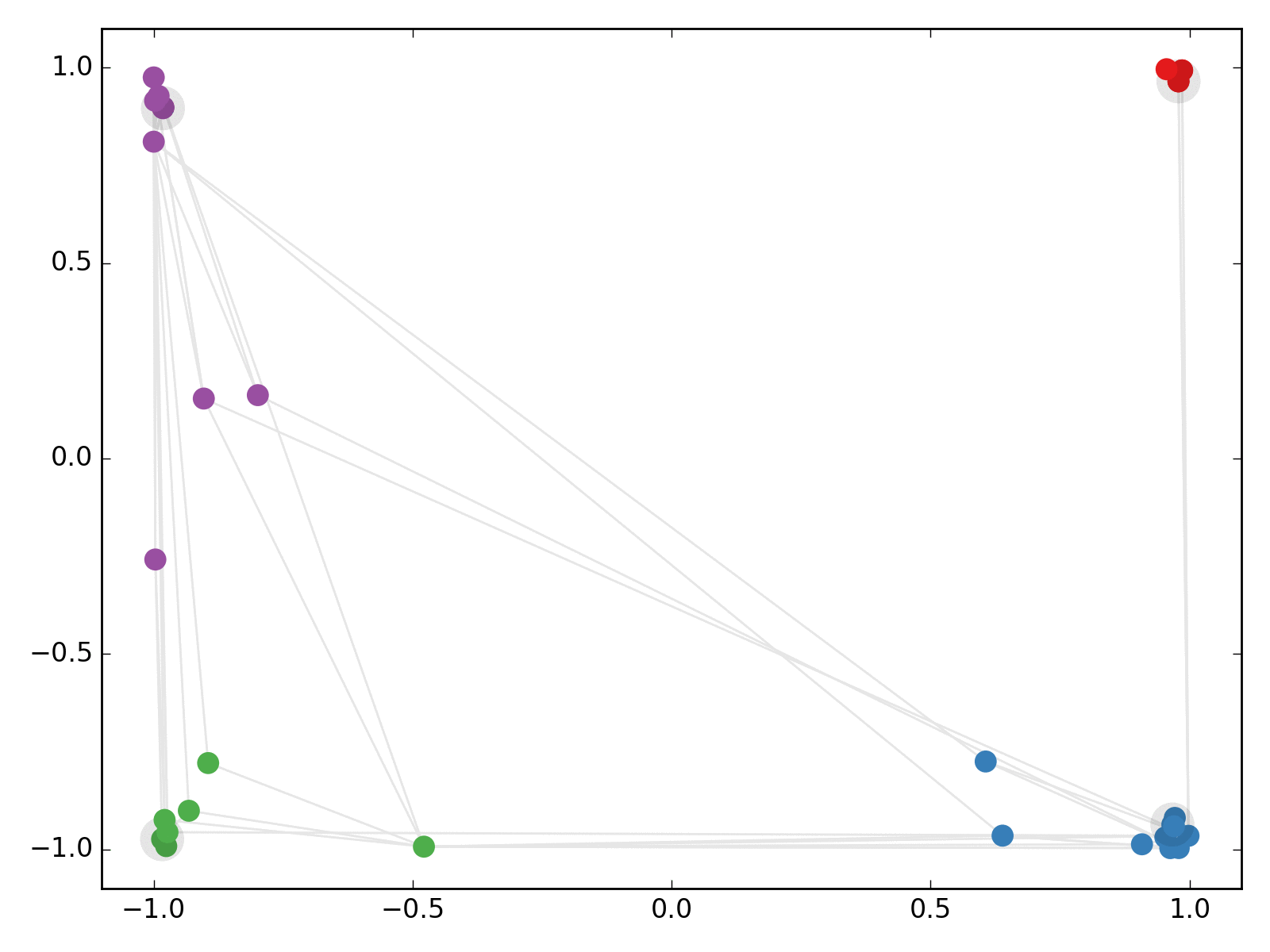}
    \caption{Iteration 200}
    \label{fig:semi-emb-e}
\end{subfigure}%
~
\begin{subfigure}[b]{0.5\textwidth}
    \centering
    \includegraphics[width=\textwidth, trim={0 1.5cm 0 0}, clip]{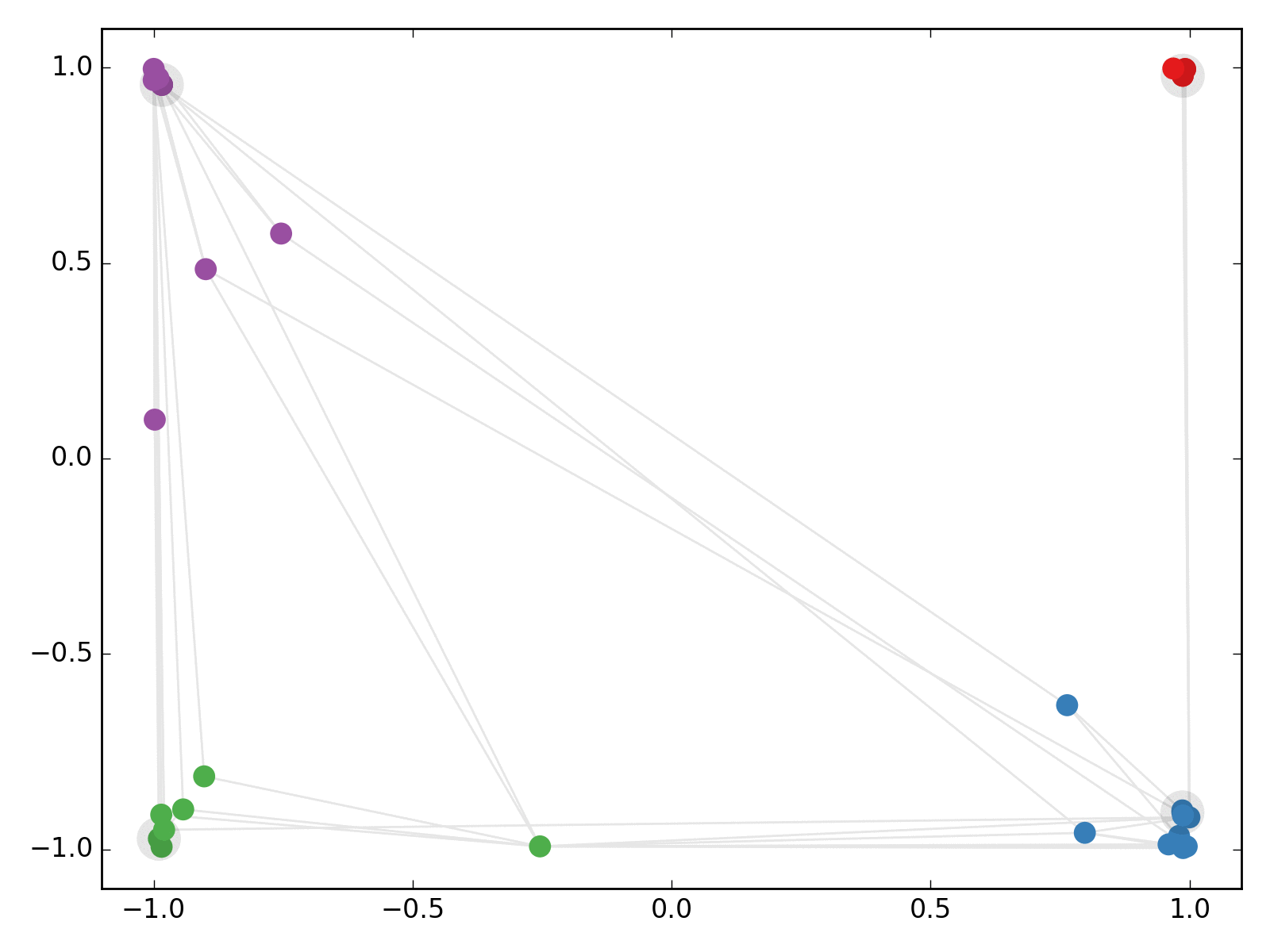}
    \caption{Iteration 300}
    \label{fig:semi-emb-f}
\end{subfigure}
\caption{Evolution of karate club network node embeddings obtained from a GCN model after a number of semi-supervised training iterations. Colors denote class. Nodes of which labels were provided during training (one per class) are highlighted (grey outline). Grey links between nodes denote graph edges. Best viewed on a computer screen.}
\label{fig:semi-emb}
\end{figure}

On this simple example of a GCN applied to the karate club network it is interesting to observe how embeddings react during training on a semi-supervised classification task. Such a visualization (see Figure \ref{fig:semi-emb}) provides insights into how the GCN model can make use of the graph structure (and of features extracted from the graph structure at later layers) to learn embeddings that are useful for a classification task.

We consider the following semi-supervised learning setup: we add a softmax layer on top of our model (\eq \ref{eq:three-layer-gcn}) and train using only a single labeled example per class (\ie a total number of 4 labeled nodes). We train for 300 training iterations using Adam \citep{kingma2014adam} with a learning rate of $0.01$ on a cross-entropy loss.

Figure \ref{fig:semi-emb} shows the evolution of node embeddings over a number of training iterations. The model succeeds in linearly separating the communities based on minimal supervision and the graph structure alone. A video of the full training process can be found on our website\footnote{\url{http://tkipf.github.io/graph-convolutional-networks/}}.

\newpage
\section{Experiments on Model Depth}
\label{sec:depth}
In these experiments, we investigate the influence of model depth (number of layers) on classification performance. We report results on a 5-fold cross-validation experiment on the Cora, Citeseer and Pubmed datasets \citep{sen2008collective} using all labels. In addition to the standard GCN model (\eq \ref{eq:gcn-layer}), we report results on a model variant where we use residual connections \citep{he2015deep} between hidden layers to facilitate training of deeper models by enabling the model to carry over information from the previous layer's input:
\begin{equation}
  \textstyle
  H^{(l+1)}= \sigma\!\left(\tilde{D}^{-\frac{1}{2}} \tilde{A}\tilde{D}^{-\frac{1}{2}}H^{(l)} W^{(l)} \right) + H^{(l)} \, .
\label{eq:gcn-residual-layer}
\end{equation}

\begin{figure}[htbp]
\centering
\begin{subfigure}[b]{0.33\textwidth}
    \centering
    \includegraphics[width=\textwidth]{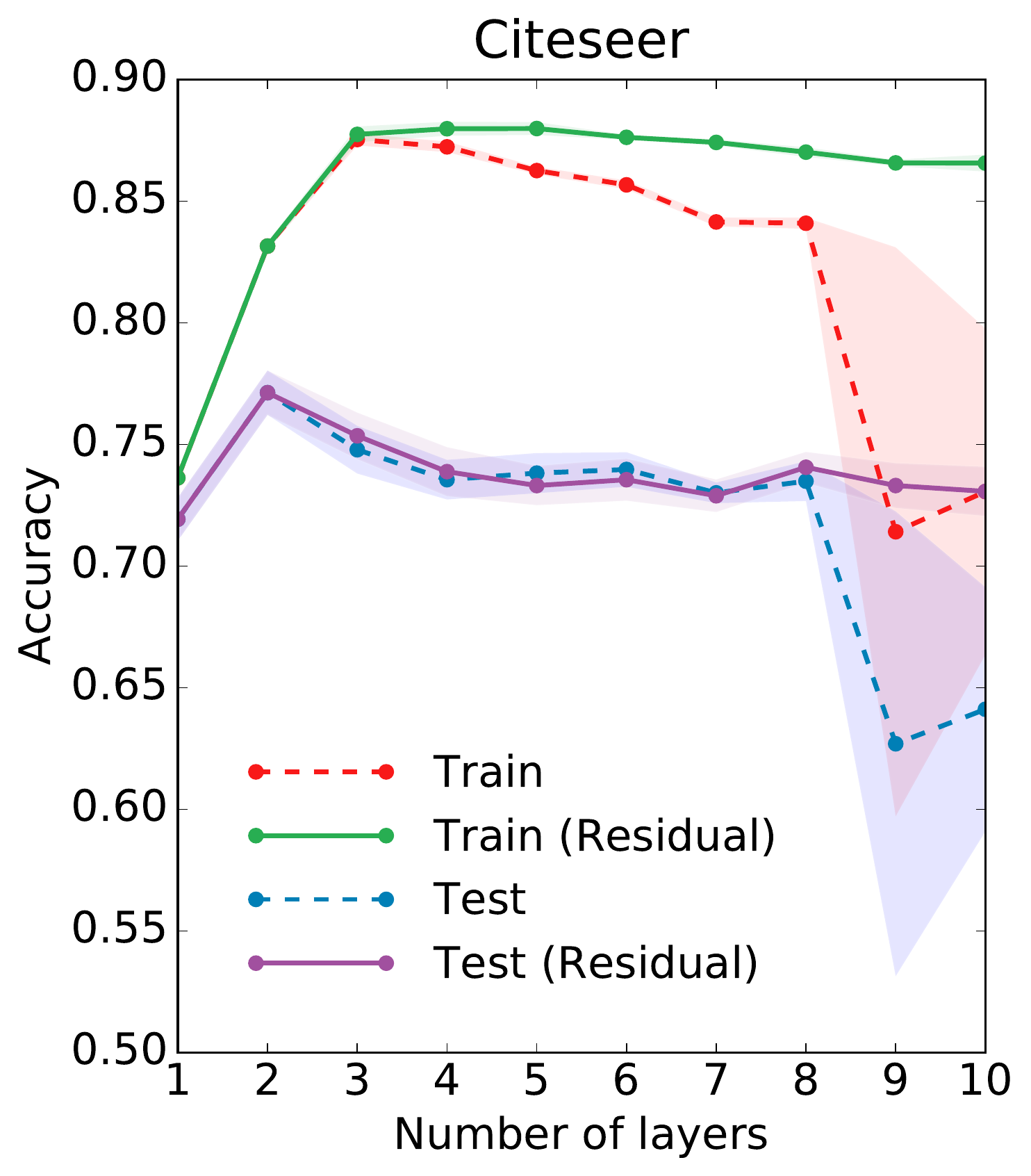}
    \label{fig:model-depth-a}
\end{subfigure}%
~
\begin{subfigure}[b]{0.33\textwidth}
    \centering
    \includegraphics[width=\textwidth]{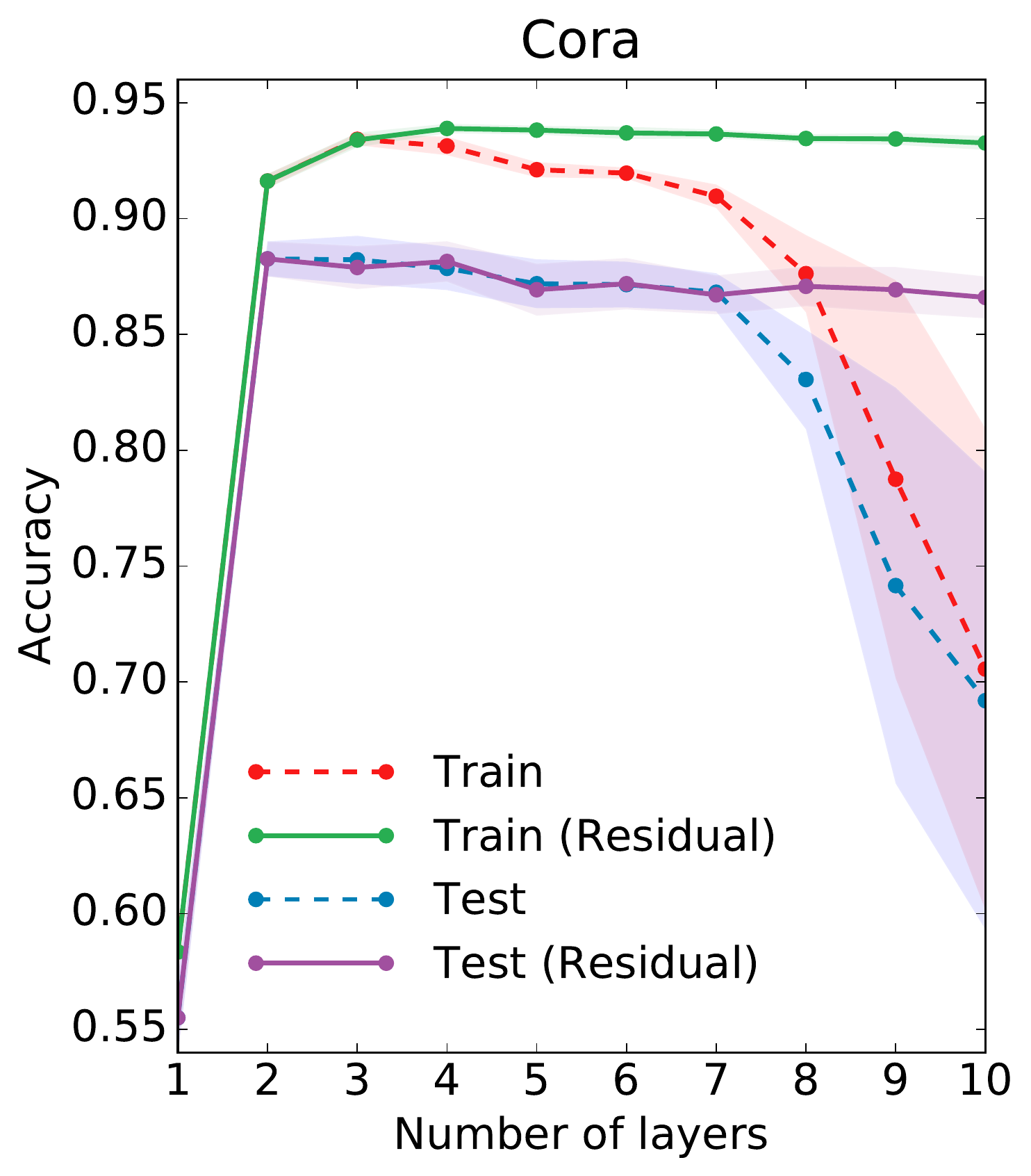}
    \label{fig:model-depth-b}
\end{subfigure}%
~
\begin{subfigure}[b]{0.33\textwidth}
    \centering
    \includegraphics[width=\textwidth]{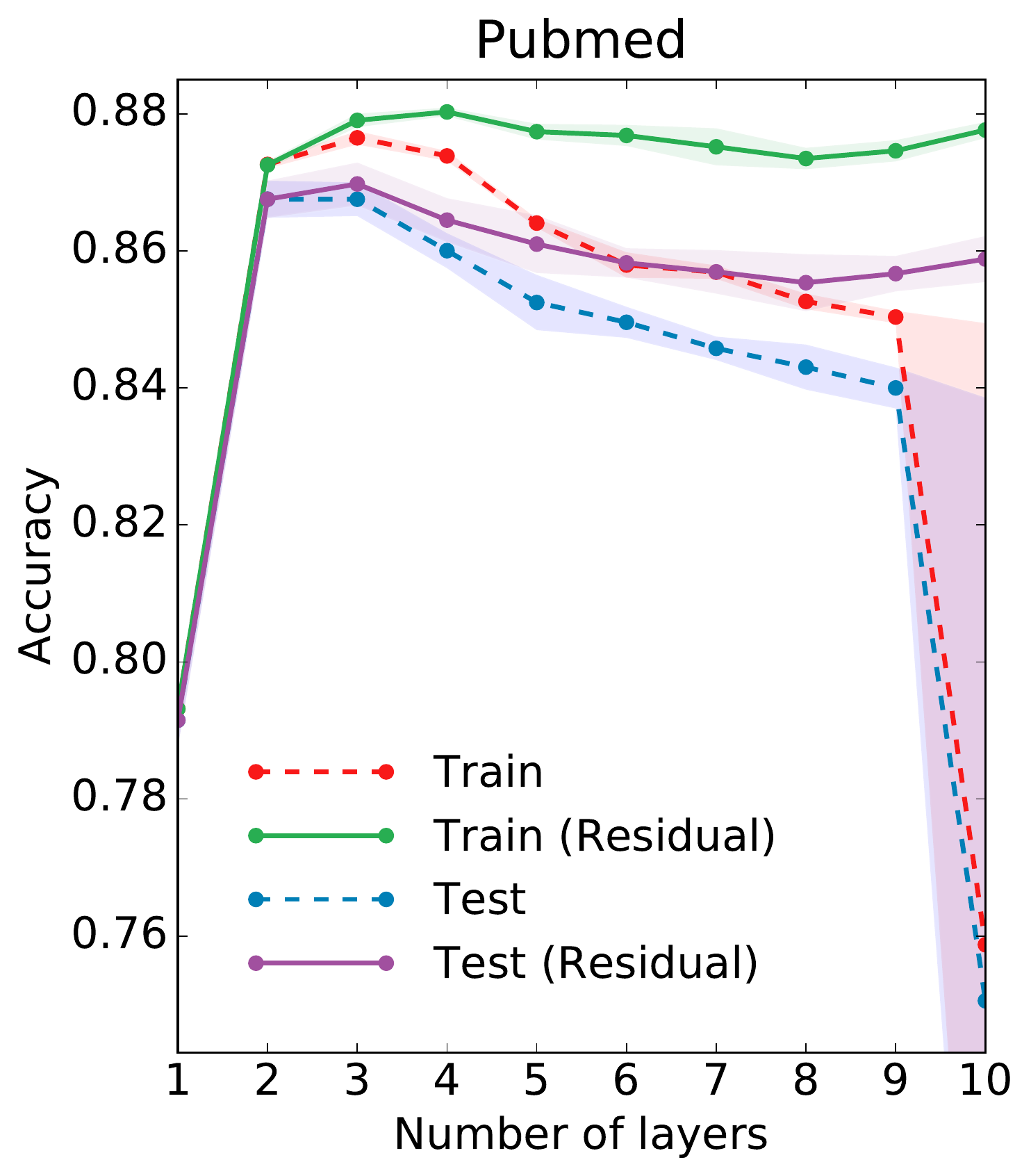}
    \label{fig:model-depth-c}
\end{subfigure}%
\vspace{-1em}
\caption{Influence of model depth (number of layers) on classification performance. Markers denote mean classification accuracy (training \vs testing) for 5-fold cross-validation. Shaded areas denote standard error. We show results both for a standard GCN model (dashed lines) and a model with added residual connections \citep{he2015deep} between hidden layers (solid lines).}
\label{fig:model-depth}
\end{figure}

On each cross-validation split, we train for 400 epochs (without early stopping) using the Adam optimizer \citep{kingma2014adam} with a learning rate of $0.01$. Other hyperparameters are chosen as follows: 0.5 (dropout rate, first and last layer), $5\cdot 10^{-4}$ (L2 regularization, first layer), 16 (number of units for each hidden layer) and 0.01 (learning rate). Results are summarized in Figure \ref{fig:model-depth}.

For the datasets considered here, best results are obtained with a 2- or 3-layer model. We observe that for models deeper than 7 layers, training without the use of residual connections can become difficult, as the effective context size for each node increases by the size of its $K^{\text{th}}$-order neighborhood (for a model with $K$ layers) with each additional layer. Furthermore, overfitting can become an issue as the number of parameters increases with model depth. 

\end{document}